\documentclass[10pt,twocolumn,letterpaper]{article}

\usepackage{iccv,times,graphicx,tabulary,multirow,xspace,xcolor}
\usepackage{amsmath,amssymb,xfrac,soul,adjustbox,setspace}
\usepackage[caption=false]{subfig}
\usepackage[british,american]{babel}
\usepackage[pagebackref=true,breaklinks=true,letterpaper=true,colorlinks,
  citecolor=citecolor,bookmarks=false]{hyperref}
\definecolor{citecolor}{RGB}{34,139,34}
\def\iccvPaperID{}\iccvfinalcopy
\ificcvfinal\fi 

\newcommand{\deemph}[1]{\textcolor{gray}{#1}}
\newcommand{\tmask}[0]{\text{TensorMask}\xspace}
\newcommand{\tpyramid}[0]{\text{tensor bipyramid}\xspace}
\newcommand{\Tpyramid}[0]{\text{Tensor bipyramid}\xspace}
\newcommand{\TPyramid}[0]{\text{Tensor Bipyramid}\xspace}
\newcommand{\trainset}[0]{\texttt{train2017}\xspace}
\newcommand{\valset}[0]{\texttt{val2017}\xspace}
\newcommand{\testset}[0]{\texttt{test-dev}\xspace}
\newcolumntype{x}[1]{>{\centering\arraybackslash}p{#1pt}}

\newlength\savewidth\newcommand\shline{\noalign{\global\savewidth\arrayrulewidth
  \global\arrayrulewidth 1pt}\hline\noalign{\global\arrayrulewidth\savewidth}}
\newcommand{\tablestyle}[2]{\setlength{\tabcolsep}{#1}\renewcommand{\arraystretch}{#2}\centering\footnotesize}
\makeatletter\renewcommand\paragraph{\@startsection{paragraph}{4}{\z@}
  {.5em \@plus1ex \@minus.2ex}{-.5em}{\normalfont\normalsize\bfseries}}\makeatother
\newcommand{\ourdefine}[2]{\begin{flushleft}\small\emph{\textbf{#1:} #2}\end{flushleft}}
\makeatletter\def\maketag@@@#1{\hbox{\m@th\normalfont\normalsize#1}}\makeatother
\newcommand{\sighw}{\sigma_\text{HW}}
\newcommand{\sigvu}{\sigma_\text{VU}}
\newcommand{\hatsighw}{\hat{\sigma}_\text{HW}}
\newcommand{\hatsigvu}{\hat{\sigma}_\text{VU}}
\def\x{\times}
\newcommand{\app}{\raise.17ex\hbox{$\scriptstyle\sim$}}
\newcommand{\mr}[1]{\multirow{2}{*}{\emph{#1}}}

\begin{document}
\title{TensorMask: A Foundation for Dense Object Segmentation\vspace{-3mm}}
\author{%
 Xinlei Chen \quad Ross Girshick \quad Kaiming He \quad Piotr Doll\'ar\\[2mm]
 Facebook AI Research (FAIR)}
\maketitle

\begin{abstract}
Sliding-window object detectors that generate bounding-box object predictions over a dense, regular grid have advanced rapidly and proven popular. In contrast, modern instance segmentation approaches are dominated by methods that first detect object bounding boxes, and then crop and segment these regions, as popularized by Mask R-CNN. In this work, we investigate the paradigm of dense sliding-window instance segmentation, which is surprisingly under-explored. Our core observation is that this task is fundamentally different than other dense prediction tasks such as semantic segmentation or bounding-box object detection, as the output at every spatial location is itself a geometric structure with its own spatial dimensions. To formalize this, we treat dense instance segmentation as a prediction task over 4D tensors and present a general framework called \emph{\tmask} that explicitly captures this geometry and enables novel operators on 4D tensors. We demonstrate that the tensor view leads to large gains over baselines that ignore this structure, and leads to results comparable to Mask R-CNN. These promising results suggest that \tmask can serve as a foundation for novel advances in dense mask prediction and a more complete understanding of the task. Code will be made available.
\end{abstract}

\section{Introduction}

The sliding-window paradigm---\emph{finding objects by looking in each window placed over a dense set of image locations}---is one of the earliest and most successful concepts in computer vision~\cite{Lecun94, Viola2001, Dollar2009, Felzenszwalb2010} and is naturally connected to convolutional networks~\cite{LeCun1989}. However, while today's top-performing object detectors rely on sliding window prediction to generate initial candidate regions, a \emph{refinement} stage is applied to these candidate regions to obtain more accurate predictions, as pioneered by Faster R-CNN~\cite{Ren2015} and Mask R-CNN~\cite{He2017} for bounding-box object detection and instance segmentation, respectively. This class of methods has dominated the COCO detection challenges~\cite{Lin2014}.

Recently, bounding-box object detectors which eschew the refinement step and focus on direct sliding-window prediction, as exemplified by SSD~\cite{Liu2016} and RetinaNet~\cite{Lin2017a}, have witnessed a resurgence and shown promising results. In contrast, the field has \emph{not} witnessed equivalent progress in dense sliding-window instance segmentation; there are no direct, dense approaches analogous to SSD / RetinaNet for mask prediction. Why is the dense approach thriving for box detection, yet entirely missing for instance segmentation? This is a question of fundamental scientific interest. \emph{The goal of this work is to bridge this gap and provide a foundation for exploring dense instance segmentation.}

\begin{figure}[t]\centering
\includegraphics[width=.99\linewidth]{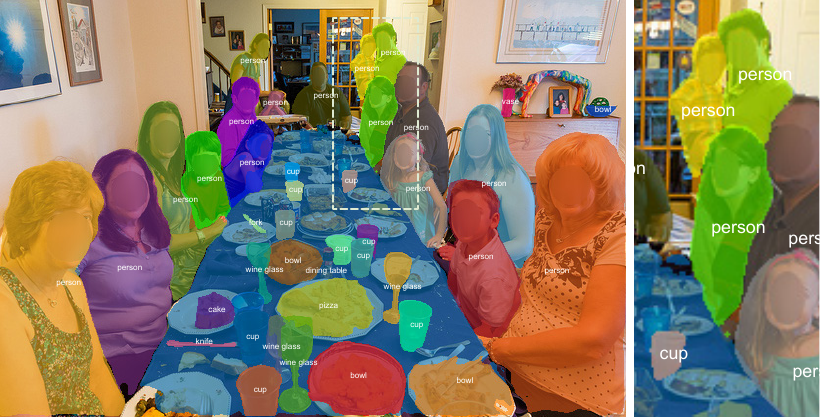}
\caption{Selected output of \emph{\tmask}, our proposed framework for performing \emph{dense sliding-window instance segmentation}. We treat dense instance segmentation as a prediction task over \emph{structured} 4D tensors. In addition to obtaining competitive quantitative results, \tmask achieves results that are \emph{qualitatively} reasonable. Observe that both small and large objects are well delineated and more critically \emph{overlapping objects are properly handled}.}
\label{fig:teaser}\vspace{-3mm}
\end{figure}

\addtocounter{footnote}{-1}
\begin{figure*}\center
 \includegraphics[width=.99\linewidth]{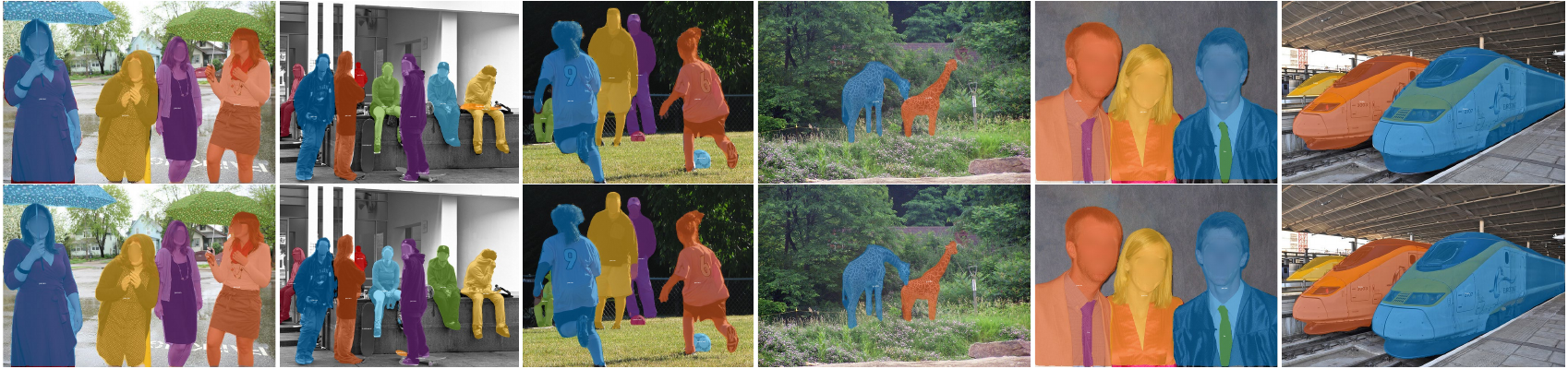}
\caption{Example results of \tmask and Mask R-CNN~\cite{He2017} with a ResNet-101-FPN backbone (on the same images as used in Fig.~6 of Mask R-CNN~\cite{He2017}). The results are \emph{quantitatively} and \emph{qualitatively} similar, demonstrating that the dense sliding window paradigm can indeed be effective for the instance segmentation task. We challenge the reader to identify which results were generated by \tmask.\protect\footnotemark}
\label{fig:qualitative}\vspace{-2mm}
\end{figure*}

Our main insight is that the core concepts for defining dense mask representations, as well as effective realizations of these concepts in neural networks, are both lacking. Unlike bounding boxes, which have a fixed, low-dimensional representation regardless of scale, segmentation masks can benefit from richer, more structured representations. For example, each mask is itself a 2D spatial map, and masks for larger objects can benefit from the use of larger spatial maps. Developing effective representations for dense masks is a key step toward enabling dense instance segmentation.

To address this, we define a set of core concepts for representing masks with high-dimensional tensors that allows for the exploration of novel network architectures for dense mask prediction. We present and experiment with several such networks in order to demonstrate the merits of the proposed representations. Our framework, called \emph{\tmask}, establishes the \emph{first} dense sliding-window instance segmentation system that achieves results near to Mask R-CNN.

\footnotetext{In Fig.~\ref{fig:qualitative}, Mask R-CNN results on top; \tmask results on bottom. }

The central idea of the \tmask representation is to use \emph{structured} 4D tensors to represent masks over a spatial domain. This perspective stands in contrast to prior work on the related task of segmenting class-agnostic object proposals such as DeepMask~\cite{Pinheiro2015} and InstanceFCN~\cite{Dai2016} that used \emph{unstructured} 3D tensors, in which the mask is packed into the third `channel' axis. The channel axis, unlike the axes representing object position, does not have a clear geometric meaning and is therefore difficult to manipulate. By using a basic channel representation, one misses an opportunity to benefit from using structural arrays to represent masks as 2D entities---analogous to the difference between MLPs and ConvNets~\cite{LeCun1989} for representing 2D images.

Unlike these channel-oriented approaches, we propose to leverage 4D tensors of shape $(V, U, H, W)$, in which both $(H, W)$---\emph{representing object position}---and $(V, U)$---\emph{representing relative mask position}---are geometric sub-tensors, \ie, they have axes with well-defined units and geometric meaning \wrt the image. This shift in perspective from encoding masks in an unstructured channel axis to using structured geometric sub-tensors enables the definition of novel operations and network architectures. These networks can operate directly on the $(V,U)$ sub-tensor in geometrically meaningful ways, including coordinate transformation, up-/downscaling, and use of scale pyramids.

Enabled by the \tmask framework, we develop a pyramid structure over a scale-indexed list of 4D tensors, which we call a \emph{\tpyramid}. Analogous to a feature pyramid, which is a list of feature maps at multiple scales, a \tpyramid contains a list of 4D tensors with shapes $(2^kV, 2^kU, \frac{1}{2^k}H, \frac{1}{2^k}W)$, where $k\ge0$ indexes scale. This structure has a pyramidal shape in \emph{both} $(H,W)$ and $(V,U)$ geometric sub-tensors, but growing in \emph{opposite} directions. This natural design captures the desirable property that large objects have high-resolution masks with coarse spatial localization (large $k$) and small objects have low-resolution masks with fine spatial localization (small $k$).

We combine these components into a network backbone and training procedure closely following RetinaNet~\cite{Lin2017a} in which our dense mask predictor extends the original dense bounding box predictor. With detailed ablation experiments, we evaluate the effectiveness of the \tmask framework and show the importance of explicitly capturing the geometric structure of this task. Finally, we show \tmask yields similar results to its Mask R-CNN counterpart (see Figs.~\ref{fig:teaser} and \ref{fig:qualitative}). These promising results suggest the proposed framework can help pave the way for future research on dense sliding-window instance segmentation.

\section{Related Work}\label{sec:related}

\paragraph{Classify mask proposals.} The modern instance segmentation task was introduced by Hariharan \etal~\cite{Hariharan2014} (before being popularized by COCO~\cite{Lin2014}). In their work, the method proposed for this task involved first generating object \emph{mask proposals}~\cite{Sande2011, Arbelaez2014}, then classifying these proposals~\cite{Hariharan2014}. In earlier work, the \emph{classify-mask-proposals} methodology was used for other tasks. For example, Selective Search~\cite{Sande2011} and the original R-CNN~\cite{Girshick2014} classified mask proposals to obtain box detections and semantic segmentation results; these methods could easily be applied to instance segmentation. These early methods relied on bottom-up mask proposals computed by pre-deep-learning era methods~\cite{Sande2011, Arbelaez2014}; our work is more closely related to dense sliding-window methods for mask object proposals as pioneered by DeepMask~\cite{Pinheiro2015}. We discuss this connection shortly.

\paragraph{Detect then segment.} The now dominant paradigm for instance segmentation involves first detecting objects with a box and then segmenting each object using the box as a guide~\cite{Dai2016b, Zagoruyko2016, Li2017, He2017}. Perhaps the most successful instantiation of the \emph{detect-then-segment} methodology is Mask R-CNN~\cite{He2017}, which extended the Faster R-CNN~\cite{Ren2015} detector with a simple mask predictor. Approaches that build on Mask R-CNN~\cite{Liu2018, Peng2018, Chen2019} have dominated leaderboards of recent challenges~\cite{Lin2014, Neuhold2017, Cordts2016}. Unlike in bounding-box detection, where sliding-window~\cite{Liu2016, Redmon2017, Lin2017a} and region-based~\cite{Girshick2015, Ren2015} methods have both thrived, in the area of instance segmentation, research on dense sliding-window methods has been missing. Our work aims to close this gap.

\paragraph{Label pixels then cluster.} A third class of approaches to instance segmentation (\eg, \cite{Bai2017, Kirillov2017, Arnab2017, Liu2017a}) builds on models developed for semantic segmentation~\cite{Long2015, Chen2015}. These approaches label each image pixel with a category and some auxiliary information that a clustering algorithm can use to group pixels into object instances. These approaches benefit from improvements on semantic segmentation and natively predict higher-resolution masks for larger objects. Compared to detect-then-segment methods, \emph{label-pixels-then-cluster} methods lag behind in accuracy on popular benchmarks~\cite{Lin2014, Neuhold2017, Cordts2016}. Instead of employing fully convolutional models for \emph{dense pixel labeling}, \tmask explores the framework of building fully convolutional (\ie, dense sliding window) models for \emph{dense mask prediction}, where the output at each spatial location is itself a 2D spatial map.

\paragraph{Dense sliding window methods.} To the best of our knowledge, \emph{no prior methods exist for dense sliding-window instance segmentation}. The proposed \tmask framework is the \emph{first} such approach. The closest methods are for the related task of class-agnostic mask \emph{proposal} generation, specifically models such as DeepMask~\cite{Pinheiro2015, Pinheiro2016} and InstanceFCN~\cite{Dai2016} which apply convolutional neural networks to generate mask proposals in a \emph{dense sliding-window} manner. Like these approaches, \tmask is a dense sliding-window model, but it spans a more expressive design space. DeepMask and InstanceFCN can be expressed naturally as class-agnostic \tmask models, but \tmask enables novel architectures that perform better. Also, unlike these class-agnostic methods, \tmask performs multi-class classification in parallel to mask prediction, and thus can be applied to the task of instance segmentation.

\section{Tensor Representations for Masks}

The central idea of the \tmask framework is to use \emph{structured high-dimensional tensors} to represent image content (\eg, masks) in a set of densely sliding windows.

Consider a $V{\x}U$ window sliding on a feature map of width $W$ and height $H$. It is possible to represent all masks in all sliding window locations by a tensor of a shape $(C, H, W)$, where each mask is parameterized by $C{=}V{\cdot}U$ pixels. This is the representation used in DeepMask~\cite{Pinheiro2015}.

The underlying spirit of this representation, however, is in fact a higher dimensional (4D) tensor with shape $(V, U, H, W)$. The sub-tensor $(V, U)$ represents a mask as a 2D spatial entity. Instead of viewing the channel dimension $C$ as a black box into which a $V{\x}U$ mask is arranged, the tensor perspective enables several important concepts for representing dense masks, discussed next.

\subsection{Unit of Length}

The \emph{unit of length} (or simply \emph{unit}) of each spatial axis is a necessary concept for understanding 4D tensors in our framework. Intuitively, the unit of an axis defines the length of one pixel along it. Different axes can have different units.

The unit of the $H$ and $W$ axes, denoted as $\sighw$, can be set as the \emph{stride} \wrt the input image (\eg, res$_4$ of ResNet-50~\cite{He2016} has $\sighw{=}16$ image pixels). Analogously, the $V$ and $U$ axes define another 2D spatial domain and have their own unit, denoted as $\sigvu$. Shifting one pixel along the $V$ or $U$ axis corresponds to shifting $\sigvu$ pixels on the input image. The unit $\sigvu$ need not be equal to the unit $\sighw$, a property that our models will benefit from.

Defining units is necessary because the interpretation of the tensor shape $(V, U, H, W)$ is ambiguous if units are not specified. For example, $(V, U)$ represents a $V{\x}U$ window in image pixels if $\sigvu{=}1$ image pixel, but a $2V{\x}2U$ window in image pixels if $\sigvu{=}2$ image pixels. The units and how they change due to up/down-scaling operations are central to multi-scale representations (more in \S\ref{sec:bipyramid}).

\subsection{Natural Representation}
\label{sec:natural}

With the definition of units, we can formally describe the representational meaning of a $(V, U, H, W)$ tensor. In our simplest definition, this tensor represents the windows sliding over $(H, W)$. We call this the \emph{natural representation}. Denoting $\alpha{=}\sfrac{\sigvu}{\sighw}$ as the ratio of units, formally we have:
\ourdefine{Natural Representation}{For a 4D tensor of shape $(V, U, H, W)$, its value at coordinates $(v, u, y, x)$ represents the mask value at $(y+ \alpha v, x+ \alpha u)$ in the $\alpha V{\x}\alpha U$ window centered at $(y, x)$.\protect\footnotemark}
Here $(v, u, y, x) \in [-\frac{V}{2}, \frac{V}{2}) {\x} [-\frac{U}{2}, \frac{U}{2}) {\x} [0, H) {\x} [0, W)$, where `$\x$' denotes cartesian product. Conceptually, the tensor can be thought of as a continuous function in this domain. For implementation, we must instead rasterize the 4D tensor as a discrete function defined on sampled locations. We assume a sampling rate of one sample per unit, with samples located at integer coordinates (\eg, if $U{=}3$, then $u {\in} \{-1,0,1\}$). This assumption allows the same value $U$ to represent both the length of the axis in terms of units (\eg, $3\sigvu$) and also the number of discrete samples stored for the axis. This is convenient for working with tensors produced by neural networks that are discrete \emph{and} have lengths.

\footnotetext{Derivation: on the input image pixels, the center of a sliding window is $(y {\cdot} \sighw, x {\cdot} \sighw)$, and a pixel located \wrt this window is at $(y {\cdot} \sighw + v {\cdot} \sigvu, x {\cdot} \sighw + u {\cdot} \sigvu)$. Projecting to the $HW$ domain (\ie, normalizing by the unit $\sighw$) gives us $(y, x)$ and $(y+ \alpha v, x+ \alpha u)$.}

\begin{figure}[t]\centering
\includegraphics[width=.99\linewidth]{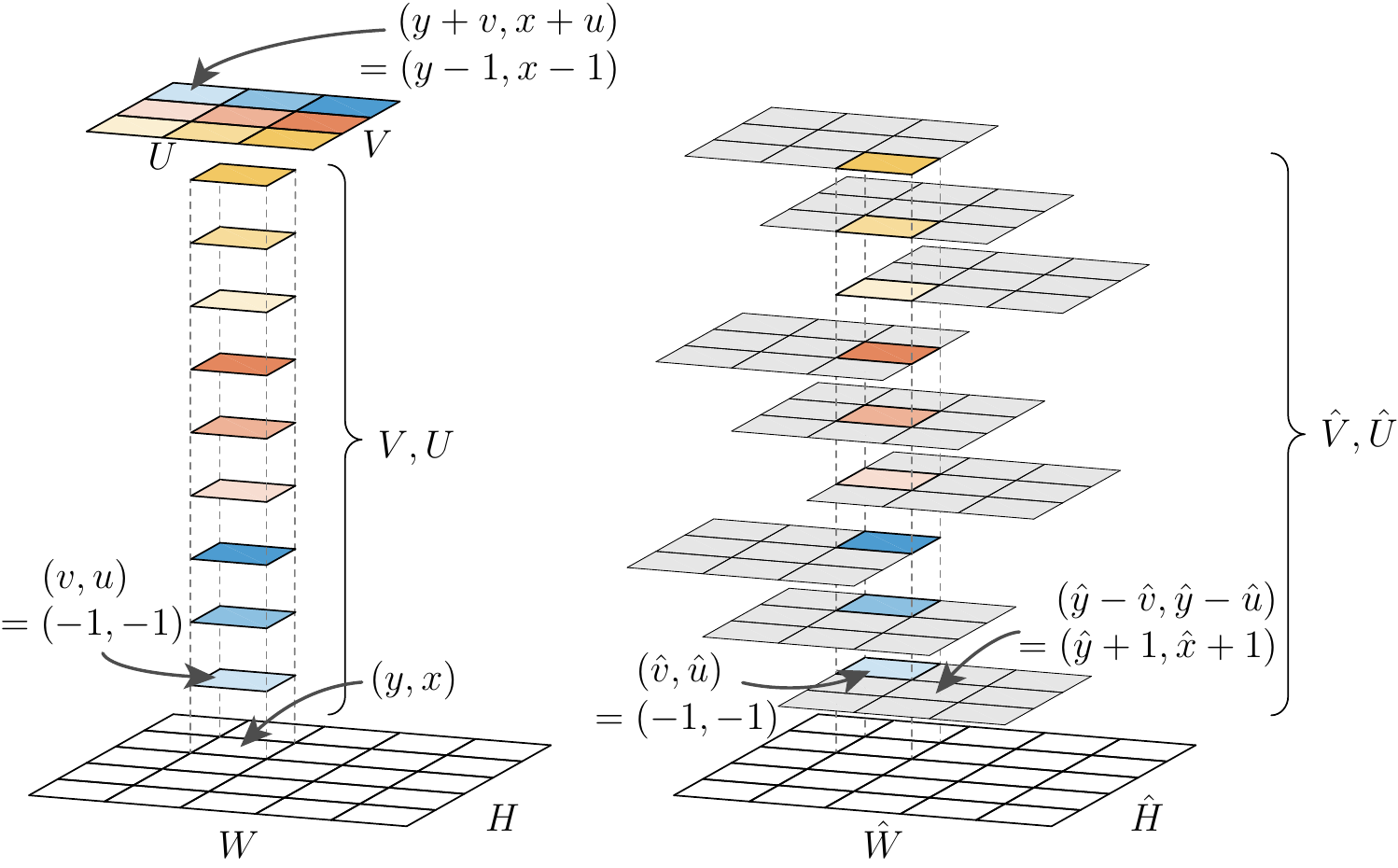}
\caption{Left: \textbf{Natural representation}. The $(V, U)$ sub-tensor at a pixel represents a window centered at this pixel. Right: \textbf{Aligned representation}. The $(\hat{V}, \hat{U})$ sub-tensor at a pixel represents the values at this pixel in each of the windows overlapping it.}
\label{fig:representations}
\end{figure}

Fig.~\ref{fig:representations} (left) illustrates an example when $V{=}U{=}3$ and $\alpha$ is 1. The natural representation is intuitive and easy to parse as the output of a network, but it is not the only possible representation in a deep network, as discussed next.

\subsection{Aligned Representation}
\label{sec:aligned}

In the natural representation, a sub-tensor $(V, U)$ located at $(y, x)$ represents values at \emph{offset} pixels $(y{+}\alpha v, x{+}\alpha u)$ instead of directly at $(y, x)$. When using convolutions to compute features, preserving \emph{pixel-to-pixel alignment} between input pixels and predicted output pixels can lead to improvements (this is similar to the motivation for RoIAlign~\cite{He2017}). Next we describe a pixel-aligned representation for dense masks under the tensor perspective.

Formally, we define the \emph{aligned representation} as:
\ourdefine{Aligned Representation}{For a 4D tensor $(\hat{V}, \hat{U}, \hat{H}, \hat{W})$, its value at coordinates $(\hat{v}, \hat{u}, \hat{y}, \hat{x})$ represents the mask value at $(\hat{y}, \hat{x})$ in the $\hat{\alpha}\hat{V}{\x}\hat{\alpha}\hat{U}$ window centered at $(\hat{y}- \hat{\alpha} \hat{v}, \hat{x}- \hat{\alpha} \hat{u})$.}
$\hat{\alpha}{=}\sfrac{\hatsigvu}{\hatsighw}$ is the ratio of units in the aligned representation.

Here, the sub-tensor $(\hat{V}, \hat{U})$ at pixel $(\hat{y}, \hat{x})$ always describes the values taken at this pixel, \ie it is \emph{aligned}. The subspace $(\hat{V}, \hat{U})$ does \emph{not} represent a single mask, but instead enumerates mask values in all $\hat{V}{\cdot}\hat{U}$ windows that overlap pixel $(\hat{y}, \hat{x})$. Fig.~\ref{fig:representations} (right) illustrates an example when $\hat{V}{=}\hat{U}{=}3$ (nine overlapping windows) and $\hat{\alpha}$ is 1.

Note that we denote tensors in the aligned representation as $(\hat{V}, \hat{U}, \hat{H}, \hat{W})$ (and likewise for coordinates/units). This is in the spirit of `\emph{named tensors}'~\cite{Rush2019} and proves useful.

Our aligned representation is related to the \emph{instance-sensitive score maps} proposed in InstanceFCN~\cite{Dai2016}. We prove (in \S\ref{sec:app:instancefcn}) that those score maps behave like our aligned representation but with nearest-neighbor interpolation on $(\hat{V},\hat{U})$, which makes them \emph{unaligned}. We test this experimentally and show it degrades results severely.

\subsection{Coordinate Transformation}
\label{sec:coordtrans}

We introduce a coordinate transformation between natural and aligned representations, so they can be used interchangeably in a single network. This gives us additional flexibility in the design of novel network architectures.

For simplicity, we assume units in both representations are the same: \ie, $\sighw{=}\hatsighw$ and $\sigvu{=}\hatsigvu$, and thus $\alpha{=}\hat{\alpha}$ (for the more general case see \S\ref{sec:app:general}). Comparing the definitions of natural \vs aligned representations, we have the following two relations for $x, u$: $x{+}\alpha u{=}\hat{x}$ and $x{=}\hat{x}{-}\hat{\alpha} \hat{u}$. With $\alpha{=}\hat{\alpha}$, solving this equation for $\hat{x}$ and $\hat{u}$ gives: $\hat{x}{=}x{+}\alpha u$ and $\hat{u}{=}u$. A similar results hold for $y, v$. So the transformation from the aligned representation ($\hat{\mathcal{F}}$) to the natural representation ($\mathcal{F}$) is:
\begin{equation}
 \mathcal{F}(v,u,y,x)=\hat{\mathcal{F}}(v, u, y + \alpha v, x + \alpha u)\label{eq:align2nat}.
\end{equation}
We call this transform \texttt{align2nat}. Likewise, solving this set of two relations for $x$ and $u$ gives the reverse transform of \texttt{nat2align}: $\hat{\mathcal{F}}(\hat{v},\hat{u},\hat{y},\hat{x}){=}\mathcal{F}(\hat{v}, \hat{u}, \hat{y}{-}\alpha\hat{v}, \hat{x}{-}\alpha\hat{u})$. While all the models presented in this work only use \texttt{align2nat}, we present both cases for completeness.

Without restrictions on $\alpha$, these transformations may involve indexing a tensor at a non-integer coordinate, \eg if $x{+}\alpha u$ is not an integer. Since we only permit integer coordinates in our implementation, we adopt a simple strategy: when the op \texttt{align2nat} is called, we ensure that $\alpha$ is a positive integer. We can satisfy this constraint on $\alpha$ by changing units with up/down-scaling ops, as described next.

\subsection{Upscaling Transformation}

\begin{figure}[t]\centering
 \includegraphics[width=.9\linewidth]{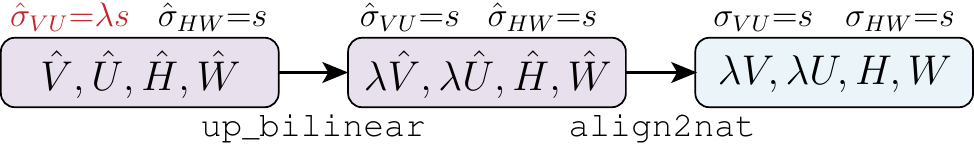}
 \caption{The \textbf{\texttt{up\_align2nat} op} is defined as a sequence of two ops. It takes an input tensor that has a coarse, $\lambda{\x}$ lower resolution on $\hat{V}\hat{U}$ (so the unit $\hatsigvu$ is $\lambda{\x}$ larger). The op performs upsampling on $\hat{V}\hat{U}$ by $\lambda$ followed by \texttt{align2nat}, resulting in an output where $\sigvu{=}\sighw{=}s$ (where $s$ is the stride).}
\label{fig:up_align2nat}\vspace{-3mm}
\end{figure}

The aligned representation enables the use of a \emph{coarse} $(\hat{V}, \hat{U})$ sub-tensors to create finer $(V, U)$ sub-tensors, which proves quite useful. Fig.~\ref{fig:up_align2nat} illustrates this transformation, which we call \texttt{up\_align2nat} and describe next.

The \texttt{up\_align2nat} op accepts a $(\hat{V}, \hat{U}, \hat{H}, \hat{W})$ tensor as input. The $(\hat{V}, \hat{U})$ sub-tensor is $\lambda{\x}$ coarser than the desired output (so its unit is $\lambda{\x}$ bigger). It performs bilinear upsampling, $\texttt{up\_bilinear}$, in the $(\hat{V}, \hat{U})$ domain by $\lambda$, reducing the underlying unit by $\lambda{\x}$. Next, the \texttt{align2nat} op converts the output into the natural representation. The full \texttt{up\_align2nat} op is shown in Fig.~\ref{fig:up_align2nat}.

As our experiments demonstrate, the \texttt{up\_align2nat} op is effective for generating high-resolution masks without inflating channel counts in preceding feature maps. This in turn enables novel architectures, as described next.

\subsection{\TPyramid}
\label{sec:bipyramid}

In multi-scale box detection it is common practice to use a \emph{lower}-resolution feature map to extract \emph{larger}-scale objects~\cite{Felzenszwalb2010, Lin2017}---this is because a sliding window of a \emph{fixed size} on a lower-resolution map corresponds to a larger region in the input image. This also holds for multi-scale mask detection. However, unlike a box that is always represented by four numbers regardless of its scale, a mask's pixel size must scale with object size in order to maintain constant resolution density. Thus, instead of always using $V{\x}U$ units to present masks of different scales, we propose to adapt the number of mask pixels based on the scale.

Consider the natural representation $(V, U, H, W)$ on a feature map of the \emph{finest} level. Here, the $(H, W)$ domain has the highest resolution (smallest unit). We expect this level to handle the \emph{smallest} objects, so the $(V, U)$ domain should have the lowest resolution. With reference to this, we build a pyramid that gradually \emph{reduces} $(H, W)$ and \emph{increases} $(V, U)$. Formally, we define a \emph{\tpyramid} as:
\ourdefine{\normalsize\Tpyramid}{\normalsize A \tpyramid is a list of tensors of shapes: $(2^k V, 2^k U, \frac{1}{2^k} H, \frac{1}{2^k} W)$, for $k{=}0, 1, 2, \ldots$, with units $\sigvu^{k+1} = \sigvu^k$ and $\sighw^{k+1} = 2\sighw^k, \forall k$.}

Because the units $\sigvu^k$ are the same across all levels, a $2^k V{\x}2^k U$ mask has $4^{k}{\x}$ more pixels in the input image. In the $(H, W)$ domain, because the units $\sighw^k$ increase with $k$, the number of predicted masks decreases for larger masks, as desired. Note that the total size of each level is the same (it is $V{\cdot} U{\cdot} H{\cdot} W$). A \tpyramid can be constructed using the \texttt{swap\_align2nat} operation, described next.

\begin{figure}[t]\centering
 \includegraphics[width=.9\linewidth]{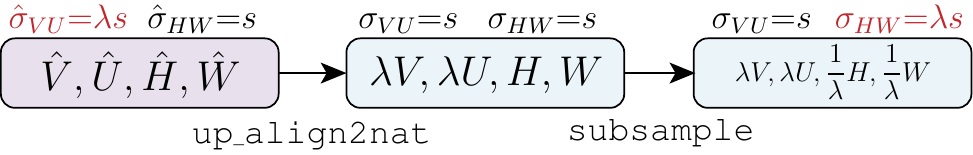}
\caption{The \textbf{\texttt{swap\_align2nat} op} is defined by two ops. It upscales the input by \texttt{up\_align2nat} (Fig.~\ref{fig:up_align2nat}), then performs \texttt{subsample} on the $HW$ domain. Note how the op \emph{swaps} the units between the $VU$ and $HW$ domains. In practice, we implement this op in place so the complexity is independent of $\lambda$.}
\label{fig:swap_align2nat}\vspace{1mm}
\end{figure}

This \texttt{swap\_align2nat} op is composed of two steps: first, an input tensor with fine $(\hat{H}, \hat{W})$ and coarse $(\hat{V}, \hat{U})$ is upscaled to $(2^k V, 2^kU, H, W)$ using \texttt{up\_align2nat}. Then $(H, W)$ is subsampled to obtain the final shape. The combination of \texttt{up\_align2nat} and \texttt{subsample}, shown in Fig.~\ref{fig:swap_align2nat}, is called \texttt{swap\_align2nat}: the units before and after this op are \emph{swapped}. For efficiency, it is not necessary to compute the intermediate tensor of shape $(2^kV, 2^kU, H, W)$ from \texttt{up\_align2nat}, which would be prohibitive. This is because only a small subset of values in this intermediate tensor appear in the final output after subsampling. So although Fig.~\ref{fig:swap_align2nat} shows the conceptual computation, in practice we implement \texttt{swap\_align2nat} as a single op that only performs the necessary computation and has complexity $O(V{\cdot} U{\cdot} H{\cdot} W)$ regardless of $k$.

\section{\tmask Architecture}
\label{sec:instantiation}

We now present models enabled by \tmask representations. These models have a mask prediction head that generates masks in sliding windows and a classification head to predict object categories, analogous to the box regression and classification heads in sliding-window object detectors~\cite{Liu2016,Lin2017a}. Box prediction is not necessary for \tmask models, but can easily be included.

\begin{figure}[t]\centering
\includegraphics[width=1\linewidth]{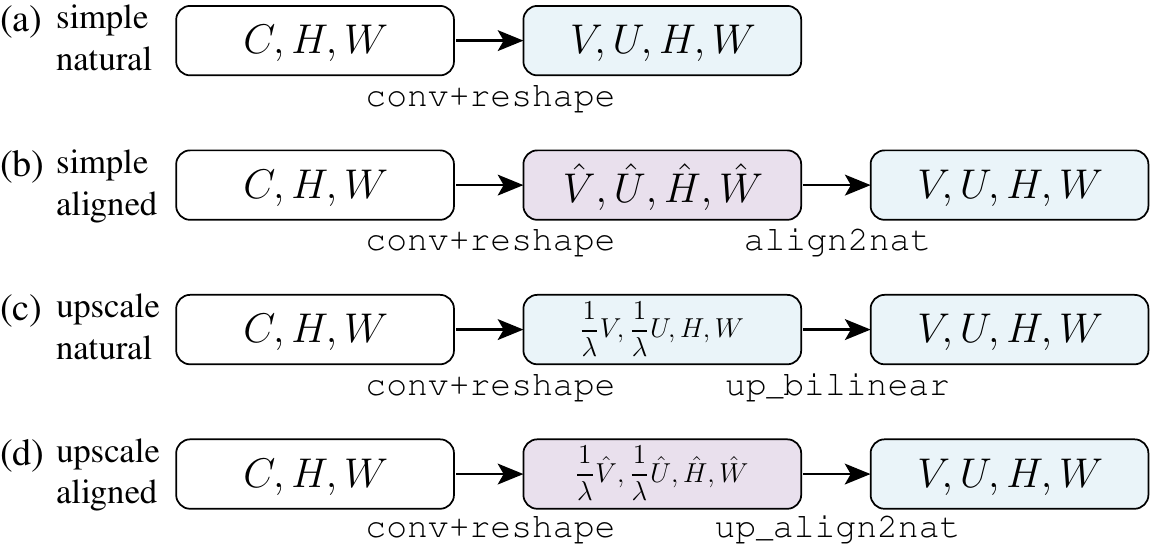}
\caption{\textbf{Baseline mask prediction heads}: Each of the four heads shown starts from a feature map (\eg, from a level of an FPN~\cite{Lin2017}) with an arbitrary channel number $C$. Then a 1$\x$1 \texttt{conv} layer projects the features into an appropriate number of channels, which form the specified 4D tensor by \texttt{reshape}. The output units of these four heads are the same, and $\sigvu{=}\sighw$.}
\label{fig:baseline_heads}
\end{figure}

\subsection{Mask Prediction Heads}
\label{sec:mask_prediction_heads}

Our mask prediction branch attaches to a convolutional backbone. We use FPN~\cite{Lin2017}, which generates a pyramid of feature maps with sizes $(C, \frac{1}{2^k} H, \frac{1}{2^k} W)$ with a fixed number of channels $C$ per level $k$. These maps are used as input for each prediction head: mask, class, and box. Weights for the heads are shared across levels, but not between tasks.

\paragraph{Output representation.} We always use the \emph{natural} representation (\S\ref{sec:natural}) as the \emph{output} format of the network. Any representation (natural, aligned, \etc) can be used in the intermediate layers, but it will be transformed into the natural representation for the output. This standardization decouples the loss definition from network design, making use of different representations simpler. Also, our mask output is \emph{class-agnostic}, \ie, the window always predicts a single mask regardless of class; the class of the mask is predicted by the classification head. Class-agnostic mask prediction avoids multiplying the output size by the number of classes.

\paragraph{Baseline heads.} We consider a set of four baseline heads, illustrated in Fig.~\ref{fig:baseline_heads}. Each head accepts an input feature map of shape $(C, H, W)$ for any $(H, W)$. It then applies a 1$\x$1 convolutional layer (with ReLU) with the appropriate number of output channels such that reshaping it into a 4D tensor produces the desired shape for the next layer, denoted as `\texttt{conv+reshape}'. Fig.~\ref{fig:baseline_heads}a and \ref{fig:baseline_heads}b are \textbf{\emph{simple heads}} that use natural and aligned representations, respectively. In both cases, we use $V{\cdot} U$ output channels for the $1{\x}1$ \texttt{conv}, followed by \texttt{align2nat} in the latter case. Fig.~\ref{fig:baseline_heads}c and \ref{fig:baseline_heads}d are \textbf{\emph{upscaling heads}} that use the natural and aligned representations, respectively. Their $1{\x}1$ \texttt{conv} has $\lambda^2{\x}$ \emph{fewer} output channels than in the simple heads.

\begin{figure}[t]\centering
\includegraphics[width=.9\linewidth]{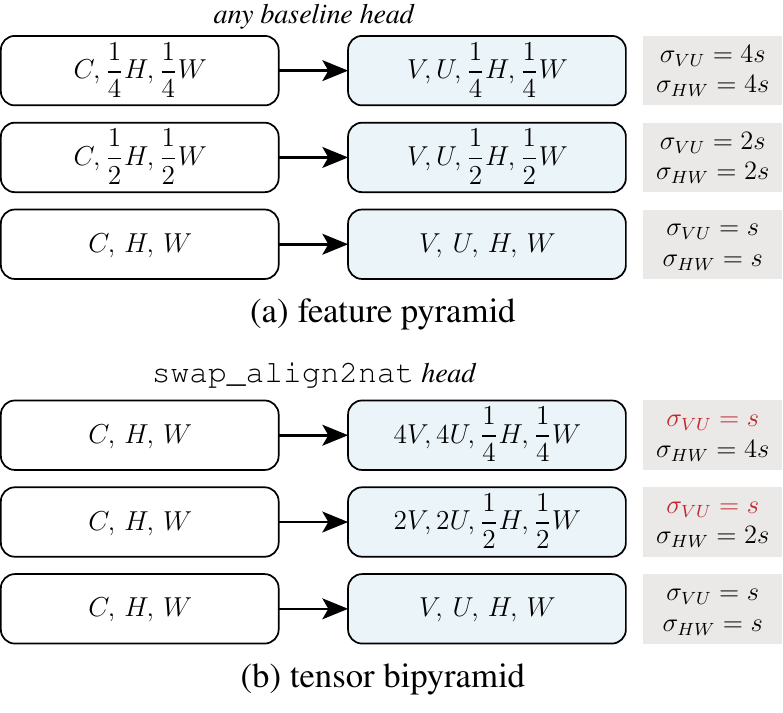}
\caption{Conceptual comparison between: (a) a \textbf{feature pyramid} with any one of the baseline heads (Fig.~\ref{fig:baseline_heads}) attached, and (b) a \textbf{\tpyramid} that uses \texttt{swap\_align2nat} (Fig.~\ref{fig:swap_align2nat}). A baseline head on the feature pyramid has $\sigvu{=}\sighw$ for each level, which implies that masks for large objects and small objects are predicted using the same number of pixels. On the other hand, the \texttt{swap\_align2nat} head can keep the mask resolution high (\ie, $\sigvu$ is the same across levels) despite the $HW$ resolution changes.}
\label{fig:pyr_bipyr}
\end{figure}

In a baseline \tmask model, one of these four heads is selected and attached to all FPN levels. The output forms a pyramid of $(V, U, \frac{1}{2^k} H, \frac{1}{2^k} W)$, see Fig.~\ref{fig:pyr_bipyr}a. For each head, the output sliding window always has the same unit as the feature map on which it slides: $\sigvu{=}\sighw$ for all levels.

\begin{figure}[t]\centering
\includegraphics[width=.9\linewidth]{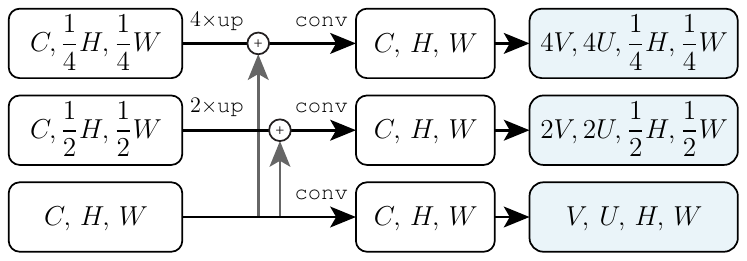}
\caption{\textbf{Conversion of FPN} feature maps from $(C, \frac{1}{2^k}H, \frac{1}{2^k}W)$ to $(C, H, W)$ for use with \tpyramid (see Fig.~\ref{fig:pyr_bipyr}b). For an FPN level $(C, \frac{1}{2^k} H, \frac{1}{2^k} W)$, we apply bilinear interpolation to upsample the feature map by a factor of $2^k$. As the upscaling can be large, we add the finest level feature map to all levels (including the finest level itself), followed by one $3{\x}3$ \texttt{conv} with ReLU.}
\label{fig:backbone}
\end{figure}

\paragraph{\Tpyramid head.} Unlike the baseline heads, the \tpyramid head (\S\ref{sec:bipyramid}) accepts a feature map of fine resolution $(H, W)$ at all levels. Fig.~\ref{fig:backbone} shows a minor modification of FPN to obtain these maps. For each of the resulting levels, now all $(C, H, W)$, we first use \texttt{conv+reshape} to produce the appropriate 4D tensor, then run a mask prediction head with \texttt{swap\_align2nat}, see Fig.~\ref{fig:pyr_bipyr}b. The \tpyramid model is the most effective \tmask variant explored in this work.

\subsection{Training}\label{sec:training}

\paragraph{Label assignment.} We use a version of the DeepMask assignment rule~\cite{Pinheiro2015} to label each window. A window satisfying three conditions \wrt a ground-truth mask $m$ is positive:

\emph{{(i)~Containment:}} the window fully contains $m$ and the longer side of $m$, in image pixels, is at least 1/2 of the longer side of the window, that is, $\max(U {\cdot} \sigvu, V {\cdot} \sigvu)$.\footnote{A fallback is used to increase small object recall: masks smaller than the minimum assignable size are assigned to windows of the smallest size.}

\emph{{(ii)~Centrality:}} the center of $m$'s bounding box is within one unit ($\sigvu$) of the window center in $\ell_2$ distance.

\emph{{(iii) Uniqueness:}} there is no other mask $m'{\ne}m$ that satisfies the other two conditions.

If $m$ satisfies these three conditions, then the window is labeled as a \emph{positive example} whose ground-truth mask, object category, and box are given by $m$. Otherwise, the window is labeled as a \emph{negative example}.

In contrast to the IoU-based assignment rules for boxes in sliding-window detectors (\eg, RPN~\cite{Ren2015}, SSD~\cite{Liu2016}, RetinaNet~\cite{Lin2017a}), our rules are \emph{mask-driven}. Experiments show that our rules work well even when using only $1$ or $2$ window sizes with a single aspect ratio of $1{:}1$, versus, \eg, RetinaNet's $9$ anchors of multiple scales and aspect ratios.

\paragraph{Loss.} For the mask prediction head, we adopt a per-pixel binary classification loss. In our setting, the ground-truth mask inside a sliding window often has a wide margin, resulting in an imbalance between foreground \vs background pixels. To address this imbalance, we set the weights for foreground pixels to $1.5$ in the binary cross-entropy loss. The mask loss of a window is averaged over all pixels in the window (note that in a \tpyramid the window size varies across levels), and the total mask loss is averaged over all positive windows (negative windows do not contribute to the mask loss).

For the classification head, we again adopt FL$^{*}$ with $\gamma{=}3$ and $\alpha{=}0.3$. For box regression, we use a parameter-free $\ell_{1}$ loss. The total loss is a weighted sum of all task losses.

\paragraph{Implementation details.} Our FPN implementation closely follows~\cite{Lin2017a}; each FPN level is output by four $3{\x}3$ \texttt{conv} layers of $C$ channels with ReLU (instead of one \texttt{conv} in the original FPN~\cite{Lin2017}). As with the heads, weights are shared across levels, but not between tasks. In addition, we found that averaging (instead of summing~\cite{Lin2017}) the top-down and lateral connections in FPN improved training stability. We use FPN levels $2$ through $7$ ($k{=}0,\ldots,5$) with $C{=}128$ channels for the four \texttt{conv} layers in the mask and box branches, and $C{=}256$ (the same as RetinaNet~\cite{Lin2017a}) for the classification branch. Unless noted, we use ResNet-50~\cite{He2016}.

For training, all models are initialized from ImageNet pre-trained weights. We use scale jitter where the shorter image side is randomly sampled from $[640,800]$ pixels~\cite{He2018}. Following SSD~\cite{Liu2016} and YOLO~\cite{Redmon2017}, which train models longer (${\app}65$ and $160$ epochs) than~\cite{Lin2017a,He2017}, we adopt the `6$\x$' schedule~\cite{He2018} (${\app}72$ epochs), which improves results. The minibatch size is $16$ images in $8$ GPUs. The base learning rate is $0.02$, with linear warm-up~\cite{Goyal2017} of 1k iterations. Other hyper-parameters are kept the same as~\cite{Detectron2018}.

\subsection{Inference}

Inference is similar to dense sliding-window object detectors. We use a single scale of $800$ pixels for the shorter image side. Our model outputs a mask prediction, a class score, and a predicted box for each sliding window. Non-maximum suppression (NMS) is applied to the top-scoring predictions using box IoU on the regressed boxes, following the settings in \cite{Lin2017}. To convert predicted soft masks to binary masks at the original image resolution, we use the same method and hyper-parameters as Mask R-CNN~\cite{He2017}.

\begin{figure}\centering
 \includegraphics[width=1\linewidth]{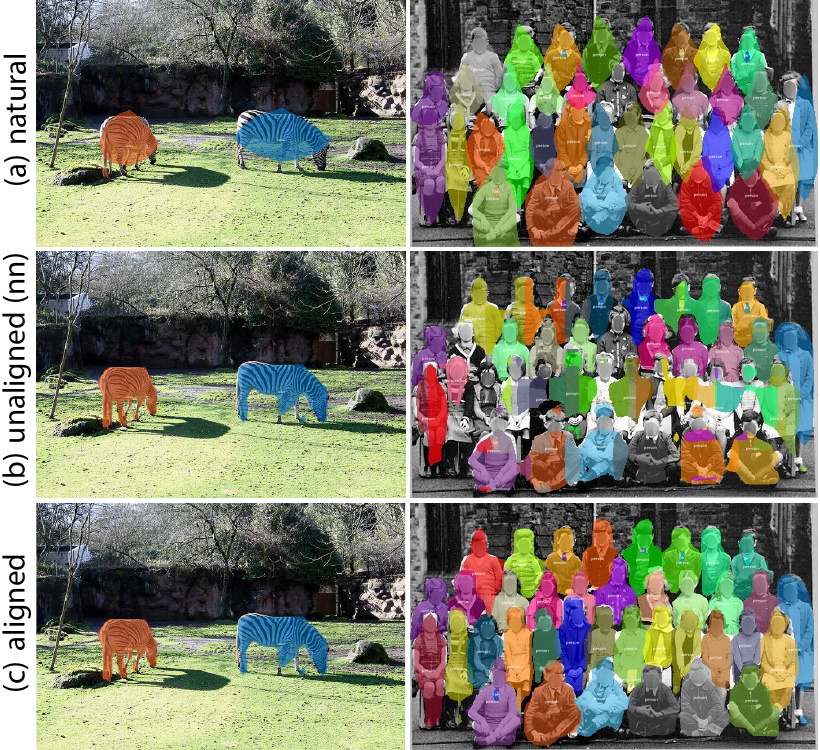}
\caption{\textbf{Baseline upscaling heads} ($\lambda{=}5$). \emph{Top:} the natural upscaling head (a) produces coarse masks, and is ineffective for large $\lambda$. \emph{Left:} for simple scenes, the unaligned head (b) and aligned head (c) (which use nearest-neighbor and bilinear interpolation, respectively), behave similarly. \emph{Right:} for overlapping objects the difference is striking: \emph{the unaligned head creates severe artifacts}.}
\label{fig:qualitative:lambda}\vspace{-1mm}
\end{figure}

\section{Experiments}
\label{sec:experiments}

We report results on COCO instance segmentation~\cite{Lin2014}. All models are trained on the $\app$118k \trainset images and tested on the 5k \valset images. Final results are on \testset. We use COCO \emph{mask} average precision (denoted by AP). When reporting \emph{box} AP, we denote it as AP$^\text{bb}$.

\begin{table}[t]
\tablestyle{7pt}{1.1}\begin{tabular}{c|x{18}x{18}x{18}|x{18}x{18}x{18}}
 head & AP & AP$_{50}$ & AP$_{75}$ & AP$_S$ & AP$_M$ & AP$_L$\\
\shline
 natural & 28.5 & 52.2 & 28.6 & 14.4 & 30.2 & 40.1 \\
 aligned & 28.9 & 52.5 & 29.3 & 14.6 & 30.8 & 40.7 \\
\end{tabular}
\vspace{2mm}\caption{\textbf{Simple heads}: natural \vs aligned (Fig.~\ref{fig:baseline_heads}a \vs \ref{fig:baseline_heads}b) with $V{\x}U{=}15{\x}15$ perform comparably if upscaling is not used.}
\label{tab:ablation:simple}\vspace{-3mm}
\end{table}

\begin{table*}[t]\centering\vspace{-5mm}
\subfloat[\textbf{Upscaling heads}: natural \vs aligned heads (Fig.~\ref{fig:baseline_heads}c \vs \ref{fig:baseline_heads}d). The $V{\x}U{=}15{\x}15$ output is upscaled by $\lambda{\x}$: \texttt{conv+reshape} uses $\frac{1}{\lambda^2}VU$ output channels as input. \emph{The aligned representation has a large gain over its natural counterpart when $\lambda$ is large}.
\label{tab:ablation:upscaling}]
{\makebox[0.475\linewidth][c]{
\tablestyle{3pt}{1.1}\begin{tabular}{c|c|x{22}x{22}x{22}|x{22}x{22}x{22}}
 head & $\lambda$ & AP & AP$_{50}$ & AP$_{75}$ & \multicolumn{3}{c}{$\Delta$ aligned - natural} \\
\shline
 natural & \multirow{2}{*}{1.5} & 28.0 & 51.7 & 27.8 & \mr{+0.9} & \mr{+0.7} & \mr{+1.5} \\
 aligned & & \textbf{28.9} & \textbf{52.4} & \textbf{29.3} & \\
\hline
 natural & \multirow{2}{*}{3} & 24.7 & 48.4 & 22.7 & \mr{+4.1} & \mr{+3.9} & \mr{+6.4} \\
 aligned & & \textbf{28.8} & \textbf{52.3} & \textbf{29.1} & \\
\hline
 natural & \multirow{2}{*}{5} & 19.2 & 42.1 & 15.6 & \mr{+9.2} & \mr{+9.7} & \mr{+13.0} \\
 aligned & & \textbf{28.4} & \textbf{51.8} & \textbf{28.6} & \\
\end{tabular}} }
\hspace{0.02\linewidth}
\subfloat[\textbf{Upscaling: bilinear \vs nearest-neighbor} interpolation for the aligned head (Fig.~\ref{fig:baseline_heads}d). The output has $V{\x}U{=}15{\x}15$. With nearest-neighbor interpolation, the aligned upscaling head is similar to the InstanceFCN~\cite{Dai2016} head.
\emph{Bilinear interpolation shows a large gain when $\lambda$ is large}.
\label{tab:ablation:upscalemethod}]
{\makebox[0.475\linewidth][c]{
\tablestyle{3pt}{1.1}\begin{tabular}{c|c|x{22}x{22}x{22}|x{22}x{22}x{22}}
 head & $\lambda$ & AP & AP$_{50}$ & AP$_{75}$ & \multicolumn{3}{c}{$\Delta$ bilinear - nearest} \\
\shline
 nearest & \multirow{2}{*}{1.5} & 28.6 & 52.1 & 29.0 & \mr{+0.3} & \mr{+0.3} & \mr{+0.3} \\
 bilinear & & \textbf{28.9} & \textbf{52.4} & \textbf{29.3} & \\
\hline
 nearest & \multirow{2}{*}{3} & 27.8 & 51.0 & 28.0 & \mr{+1.0} & \mr{+1.3} & \mr{+1.1} \\
 bilinear & & \textbf{28.8} & \textbf{52.3} & \textbf{29.1} & \\
\hline 
 nearest & \multirow{2}{*}{5} & 25.3 & 47.6 & 25.0 & \mr{+3.1} & \mr{+4.2} & \mr{+3.6} \\
 bilinear & & \textbf{28.4} & \textbf{51.8} & \textbf{28.6} & \\
\end{tabular}} }\\[-1mm]
\subfloat[
    The \textbf{\tpyramid} substantially improves results compared to the best baseline head (Tab.~\ref{tab:ablation:upscaling}, row 2) on a \emph{feature pyramid} (Fig.~\ref{fig:pyr_bipyr}a).
\label{tab:ablation:tensorpyramid}]
{\makebox[0.475\linewidth][c]{
\tablestyle{2pt}{1.1}
\begin{tabular}{c|x{22}x{22}x{22}|x{22}x{22}x{22}}
head & AP & AP$_{50}$ & AP$_{75}$ & AP$_S$ & AP$_M$ & AP$_L$ \\
\shline
 feature pyramid, best & 28.9 & 52.5 & 29.3 & 14.6 & 30.8 & 40.7 \\
 \tpyramid & \textbf{34.0} & \textbf{55.2} & \textbf{35.8} & \textbf{15.3} & \textbf{36.3} & \textbf{48.4} \\
\hline
 $\Delta$ & \emph{+5.1} & \emph{+2.7} & \emph{+6.5} & \emph{+0.7} & \emph{+5.5} & \emph{+7.7} \\
\end{tabular}} }
\hspace{0.02\linewidth}
\subfloat[\textbf{Window sizes}: extending from one $V{\x}U$ window size (per level) to two increases all AP metrics. Both rows use the \tpyramid.\label{tab:ablation:2window}]
{\makebox[0.475\linewidth][c]{
\tablestyle{2pt}{1.1}\begin{tabular}{c|x{22}x{22}x{22}|x{22}x{22}x{22}}
$V{\x}U$ & AP & AP$_{50}$ & AP$_{75}$ & AP$_S$ & AP$_M$ & AP$_L$ \\
\shline 
 $15{\x}15$ & 34.0 & 55.2 & 35.8 & 15.3 & 36.3 & 48.4 \\
 $15{\x}15,11{\x}11$ & \textbf{35.2} & \textbf{56.4} & \textbf{37.0} & \textbf{17.4} & \textbf{37.4} & \textbf{49.7} \\
\hline
$\Delta$ & \emph{+1.2} & \emph{+1.2} & \emph{+1.2} & \emph{+2.1} & \emph{+1.1} & \emph{+1.3}\\
\end{tabular}} }
\vspace{1mm}\caption{\textbf{Ablations on \tmask representations} on COCO \valset. All variants use ResNet-50-FPN and a 72 epoch schedule.\label{tab:ablation}}\vspace{2mm}
\end{table*}

\begin{table*}[t]
\tablestyle{5pt}{1.1}
\begin{tabular}{l|c|c|c|x{22}x{22}x{22}|x{22}x{22}x{22}}
 method & backbone & aug & epochs & AP & AP$_{50}$ & AP$_{75}$ & AP$_S$ & AP$_M$ & AP$_L$\\
\shline
 \deemph{Mask R-CNN~\cite{Detectron2018}} & \deemph{R-50-FPN} & & 24 & \deemph{34.9} & \deemph{57.2} & \deemph{36.9} & \deemph{15.4} & \deemph{36.6} & \deemph{50.8}\\
 Mask R-CNN, \emph{ours} & R-50-FPN & & 24 & 34.9 & 56.8 & 36.8 & 15.1 & 36.7 & 50.6 \\
 Mask R-CNN, \emph{ours} & R-50-FPN & $\checkmark$ & 72 & \textbf{36.8} & \textbf{59.2} & \textbf{39.3} & \textbf{17.1} & \textbf{38.7} & \textbf{52.1} \\
 \tmask & R-50-FPN & $\checkmark$ & 72 & 35.4 & 57.2 & 37.3 & 16.3 & 36.8 & 49.3 \\
\hline
 Mask R-CNN, \emph{ours} & R-101-FPN & $\checkmark$ & 72& \textbf{38.3} & \textbf{61.2} & \textbf{40.8} & \textbf{18.2} & \textbf{40.6} & \textbf{54.1} \\
 \tmask & R-101-FPN & $\checkmark$ & 72 & 37.1 & 59.3 & 39.4 & 17.4 & 39.1 & 51.6 \\
\end{tabular}
\vspace{1mm}\caption{\textbf{Comparison with Mask R-CNN} for instance segmentation on COCO \testset. }
\label{tab:final_mask}\vspace{-1mm}
\end{table*}

\subsection{\tmask Representations}

First we explore various tensor representations for masks using $V{=}U{=}15$ and a ResNet-50-FPN backbone. We report quantitative results in Tab.~\ref{tab:ablation} and show qualitative comparisons in Figs.~\ref{fig:qualitative} and ~\ref{fig:qualitative:lambda}.

\paragraph{Simple heads.} Tab.~\ref{tab:ablation:simple} compares natural \vs aligned representations with simple heads (Fig.~\ref{fig:baseline_heads}a \vs \ref{fig:baseline_heads}b). Both representations perform similarly, with a marginal gap of 0.4 AP. The simple natural head can be thought of as a \emph{class-specific} variant of DeepMask~\cite{Pinheiro2015} with an FPN backbone~\cite{Lin2017} and focal loss~\cite{Lin2017a}. As we aim to use lower-resolution intermediate representations, we explore upscaling heads next.

\paragraph{Upscaling heads.} Tab.~\ref{tab:ablation:upscaling} compares natural \vs aligned representations with upscaling heads (Fig.~\ref{fig:baseline_heads}c \vs \ref{fig:baseline_heads}d). The output size is fixed at $V{\x}U{=}15{\x}15$. Given an upscaling factor $\lambda$, the \texttt{conv} in Fig.~\ref{fig:baseline_heads} has $\frac{1}{\lambda^2}VU$ channels, \eg, $9$ channels with $\lambda{=}5$ (\vs $225$ channels if no upscaling). The difference in accuracy is big for large $\lambda$: the aligned variant improves AP \emph{+9.2} over the natural head (48\% relative) when $\lambda{=}5$.

The visual difference is clear in Fig.~\ref{fig:qualitative:lambda}a (natural) \vs \ref{fig:qualitative:lambda}c (aligned). The upscale aligned head still produces sharp masks with large $\lambda$. \emph{This is critical for the \tpyramid, where we have an output of $2^kV{\x}2^k U$, which is achieved with a large upscaling factor of $\lambda{=}2^k$ (\eg, $32$)}; see Fig.~\ref{fig:swap_align2nat}.

\paragraph{Interpolation.} The tensor view reveals the $(\hat{V}, \hat{U})$ sub-tensor as a 2D spatial entity that can be manipulated. Tab.~\ref{tab:ablation:upscalemethod} compares the upscale aligned head with bilinear (default) \vs \emph{nearest-neighbor} interpolation on $(\hat{V}, \hat{U})$. We refer to this latter variant as \emph{unaligned} since quantization breaks pixel-to-pixel alignment. The unaligned variant is related to InstanceFCN~\cite{Dai2016} (see \S\ref{sec:app:instancefcn}).

We observe in Tab.~\ref{tab:ablation:upscalemethod} that bilinear interpolation yields solid improvements over nearest-neighbor interpolation, especially if $\lambda$ is large ($\Delta$AP${=}$\emph{3.1}). These interpolation methods lead to striking visual differences when objects overlap: see Fig.~\ref{fig:qualitative:lambda}b (unaligned) \vs \ref{fig:qualitative:lambda}c (aligned).

\paragraph{\Tpyramid.} Replacing the best feature pyramid model with a \tpyramid yields a large 5.1 AP improvement (Tab.~\ref{tab:ablation:tensorpyramid}). Here, the mask size is $V{\x}U{=}15{\x}15$ on level $k{=}0$, and is $32V{\x}32U{=}480{\x}480$ for $k{=}5$; see Fig.~\ref{fig:pyr_bipyr}b. The higher resolution masks predicted for large objects (\eg, at $k{=}5$) have clear benefit: AP$_L$ jumps by 7.7 points. This improvement does \emph{not} come at the cost of denser windows as the $k{=}5$ output is at $(\frac{H}{32}, \frac{W}{32})$ resolution.

Again, we note that it is intractable to have, \eg, a $480^2$-channel \texttt{conv}. The upscaling aligned head with bilinear interpolation is key to making \tpyramid possible.

\paragraph{Multiple window sizes.} Thus far we have used a single window size (per-level) for all models, that is, $V{\x}U{=}15{\x}15$. Analogous to the concept of \emph{anchors} in RPN~\cite{Ren2015} that are also used in current detectors~\cite{Redmon2017,Liu2016,Lin2017a}, we extend our method to multiple window sizes. We set $V{\x}U{\in}\{15{\x}15,11{\x}11\}$, leading to two heads per level. Tab.~\ref{tab:ablation:2window} shows the benefit of having two window sizes: it increases AP by 1.2 points. More window sizes and aspect ratios are possible, suggesting room for improvement.

\subsection{Comparison with Mask R-CNN}

Tab.~\ref{tab:final_mask} summarizes the best \tmask model on \testset and compares it to the current dominant approach for COCO instance segmentation: Mask R-CNN~\cite{He2017}. We use the \texttt{Detectron}~\cite{Detectron2018} code to reflect improvements since~\cite{He2017} was published. We modify it to match our implementation details (FPN average fusion, 1k warm-up, and $\ell_{1}$ box loss). Tab.~\ref{tab:final_mask} row~1~\&~2 verify that these subtleties have a negligible effect. Then we use training-time scale augmentation and a longer schedule~\cite{He2018}, which yields an $\app$2 AP increase (Tab.~\ref{tab:final_mask}~row~3) and establishes a fair and solid baseline for comparison.

The best \tmask in Tab.~\ref{tab:ablation:2window} achieves 35.4 mask AP on \testset (Tab.~\ref{tab:final_mask} row~4), close to Mask R-CNN counterpart's 36.8. With ResNet-101, \tmask achieves 37.1 mask AP with a 1.2 AP gap behind Mask R-CNN. These results demonstrate that dense sliding-window methods \emph{can} close the gap to `detect-then-segment' systems (\S\ref{sec:related}). Qualitative results are shown in Figs.~\ref{fig:qualitative}, \ref{fig:qualitative2}, and \ref{fig:qualitative3}.

We report box AP of \tmask in \S\ref{sec:app:bbox}. Moreover, compared to Mask R-CNN, one intriguing property of \tmask is that masks are \emph{independent} from boxes. In fact, we find joint training of box and mask only gives marginal gain over mask-only training, see \S\ref{sec:app:maskonly}.

Speed-wise, the best R-101-FPN \tmask runs at 0.38s/im on a V100 GPU (all post-processing included), \vs Mask R-CNN's 0.09s/im. Predicting masks in dense sliding windows ($>$100k) results in a computation overhead, \vs Mask R-CNN's sparse prediction on $\le$100 final boxes. Accelerations are possible but outside the scope of this work.

\paragraph{Conclusion.} \tmask is a dense sliding-window instance segmentation framework that, for the \emph{first} time, achieves results close to the well-developed Mask R-CNN framework---both qualitatively and quantitatively. It establishes a conceptually complementary direction for instance segmentation research. We hope our work will create new opportunities and make both directions thrive.

\appendix
\section{Appendix}
\subsection{Generalized Coordinate Transformation}
\label{sec:app:general}

In Sec.~\ref{sec:coordtrans} we have assumed $\sighw{=}\hatsighw$ and $\sigvu{=}\hatsigvu$. Here we relax this condition and only assume $\sighw{=}\hatsighw$. Again, we still have the following two relations for $x, u$: $x{+}\alpha u{=}\hat{x}$ and $x{=}\hat{x}{-}\hat{\alpha} \hat{u}$. Solving for $\hat{x}$ and $\hat{u}$ gives: $\hat{x}{=}x{+}\alpha u$ and $\hat{u}{=}\frac{\alpha}{\hat{\alpha}}u$.
Then \texttt{align2nat} is:
\begin{eqnarray}
 \mathcal{F}(v,u,y,x)=\hat{\mathcal{F}}(\frac{\alpha}{\hat{\alpha}} v, \frac{\alpha}{\hat{\alpha}} u,y {+}\alpha v, x{+}\alpha u).
\end{eqnarray}

More generally, consider arbitrary units $\sighw$, $\hatsighw$, $\sigvu$, and $\hatsigvu$. Then the relations between the natural and aligned representation can be rewritten as:
\begin{eqnarray}
\left\{
  \begin{array}{rcl}
    x{\cdot}\sighw{+}u{\cdot}\sigvu &=& \hat{x}{\cdot}\hatsighw  \\
    x{\cdot}\sighw &=& \hat{x}{\cdot}\hatsighw{-}\hat{u}{\cdot}\hatsigvu \\
  \end{array}
\right.
\end{eqnarray}
Note that these relations only hold in the image pixel domain (hence the usage of all units). Solving for $\hat{x}$, $\hat{u}$ gives:
\begin{eqnarray}
\left\{
  \begin{array}{rcl}
    \hat{x} &=& \frac{\sighw}{\hatsighw} x{+}\frac{\sigvu}{\hatsighw} u  \\
    \hat{u} &=& \frac{\sigvu}{\hatsigvu} u \\
  \end{array}
\right.
\end{eqnarray}
And the \texttt{align2nat} transform becomes:
{\scriptsize\begin{eqnarray}
\mathcal{F}(v,u,y,x)=\hat{\mathcal{F}}(
\frac{\sigvu}{\hatsigvu} v,
\frac{\sigvu}{\hatsigvu} u,
\frac{\sighw}{\hatsighw} y{+}\frac{\sigvu}{\hatsighw} v,
\frac{\sighw}{\hatsighw} x{+}\frac{\sigvu}{\hatsighw} u).
\hspace{-2mm}\end{eqnarray}}%
This version of the coordinate transformation demonstrates the role of units and may enable more general uses.

\subsection{Aligned Representation and InstanceFCN}
\label{sec:app:instancefcn}

We prove that the InstanceFCN~\cite{Dai2016} output behaves as an upscaling aligned head with \emph{nearest-neighbor} interpolation.

In~\cite{Dai2016}, each output mask has $V{\x}U$ pixels that are divided into $K{\x}K$ bins. A mask pixel is read from the channel corresponding to the pixel's bin. In our notation, \cite{Dai2016} predicts $\mathcal{G}$ which is related to the natural representation $\mathcal{F}$ by:
\begin{align}
 \mathcal{F}(v,u,y,x) = \mathcal{G}( [ \frac{K}{V} v ], [ \frac{K}{U} u ], y + v, x + u),
 \label{eq:app:nn_derive_1}
\end{align}
where $[ \cdot ]$ is a rounding operation and the integers $[\frac{K}{V} v]$ and $[\frac{K}{U} u]$ index a bin. Now, define a new function $\tilde{\mathcal{F}}$ by:
\begin{align}
 \tilde{\mathcal{F}}(v, u, y{+}v, x{+}u)
 \triangleq \mathcal{G}( [ \frac{K}{V} v ], [ \frac{K}{U} u ], y{+}v, x{+}u),
 \label{eq:app:nn_derive_2}
\end{align}
and new coordinates: $\tilde{x}{=}x{+}u$ and $\tilde{u}{=}u$
(likewise for $v$ and $y$). Then $\tilde{\mathcal{F}}$ can be written as:
\begin{align}
 \tilde{\mathcal{F}}(\tilde{v}, \tilde{u}, \tilde{y}, \tilde{x})
 \triangleq \mathcal{G}( [ \frac{K}{V} \tilde{v} ], [ \frac{K}{U} \tilde{u} ], \tilde{y}, \tilde{x}).
 \label{eq:app:nn_derive_3}
\end{align}
Eqn.(\ref{eq:app:nn_derive_3}) says that $\mathcal{\tilde{F}}$ is the {\emph{nearest-neighbor}} interpolation of $\mathcal{G}$ on $(\tilde{V}, \tilde{U})$. Eqn.(\ref{eq:app:nn_derive_2}), (\ref{eq:app:nn_derive_1}), and the new coordinates show that $\mathcal{F}$ is computed from $\mathcal{\tilde{F}}$ by the \texttt{align2nat} transform with $\alpha{=}1$. Thus, InstanceFCN masks can be constructed in the \tmask framework by predicting $\mathcal{G}$, performing nearest-neighbor interpolation of $\mathcal{G}$ on $(\tilde{V},\tilde{U})$ to get $\mathcal{\tilde{F}}$, and then using \texttt{align2nat} to compute natural masks $\mathcal{F}$.

\subsection{Object Detection Results}
\label{sec:app:bbox}

In Tab.~\ref{tab:final_box} we show the associated \emph{bounding-box} (bb) object detection results. Overall, \tmask has a comparable box AP with Mask R-CNN and outperforms RetinaNet.

\begin{table}[h]
\tablestyle{5.5pt}{1.2}
\begin{tabular}{l|c|c|x{22}x{22}x{22}}
method & aug & epochs & AP$^\text{bb}$ & AP$^\text{bb}_{50}$ & AP$^\text{bb}_{75}$ \\
\shline
 RetinaNet, \emph{ours} & & 24 & 37.1 & 55.0 & 39.9 \\
 RetinaNet, \emph{ours} & $\checkmark$ & 72 & 39.3 & 57.2 & 42.4 \\
\hline
 Faster R-CNN, \emph{ours} & $\checkmark$ & 72 & 40.6 & 61.4 & 44.2 \\
 Mask R-CNN, \emph{ours} & $\checkmark$ & 72 & 41.7 & 62.5 & 45.7 \\
\hline
 \tmask, \emph{box-only} & $\checkmark$ & 72 & 40.8 & 60.4 & 43.9 \\
 \tmask & $\checkmark $& 72 & 41.6 & 61.0 & 45.1 \\
\end{tabular}\vspace{1mm}
\caption{\textbf{Object detection} \emph{box} AP on COCO \testset. All models use ResNet-50-FPN. `\tmask, \emph{box-only}' is our model without the mask head: it resembles RetinaNet but with the mask-driven assignment rule and only 2 window sizes instead of 9 \cite{Lin2017a}.}
\label{tab:final_box}\vspace{-2mm}
\end{table}

\subsection{Mask-Only \tmask}
\label{sec:app:maskonly}

One intriguing property of \tmask is that \emph{masks are not dependent on boxes}. This not only opens up new model designs that are mask-specific, but also allows us to investigate whether \emph{box predictions improve masks in a multi-task setting.} Here, we conduct experiments \emph{without} the use of a box head. Note that although we predict masks densely, we still need to perform NMS for post-processing. If regressed boxes are absent, we simply use the bounding boxes of the masks as a substitute (and also to report box AP).

Tab.~\ref{tab:mask_vs_box} gives the results. We observe a slight degradation switching from the default setting which uses original boxes (row 1) for NMS to using mask bounding boxes (row 2). After accounting for this, \tmask \emph{without a box head} (row 3) has nearly equal mask AP to the mask+box variant (row 2). These results indicate that the role of the box head is auxiliary in our system, in contrast to Mask R-CNN.

\begin{table}[h]
\tablestyle{3.0pt}{1.2}
\begin{tabular}{c|c|x{22}x{22}x{22}|x{22}x{22}x{22}}
\scriptsize box head & \scriptsize NMS
 & AP & AP$_{50}$ & AP$_{75}$ & AP$^\text{bb}$ & AP$^\text{bb}_{50}$ & AP$^\text{bb}_{75}$\\
\shline 
 \checkmark & bb      & 35.2 & 56.4 & 37.0 & 41.6 & 60.8 & 44.8 \\
 \checkmark & mask-bb & 34.9 & 56.0 & 36.7 & 39.7 & 59.1 & 41.8 \\
            & mask-bb & 34.8 & 56.1 & 36.7 & 39.4 & 58.8 & 41.6 \\
\end{tabular}\vspace{1mm}
\caption{\textbf{Multi-task benefits} of box training for mask prediction on COCO \valset with our final ResNet-50-FPN model.}
\label{tab:mask_vs_box}\vspace{-2mm}
\end{table}

\subsection{Qualitative Comparisons and Calibration}

\begin{figure*}\center
 \includegraphics[width=1\linewidth]{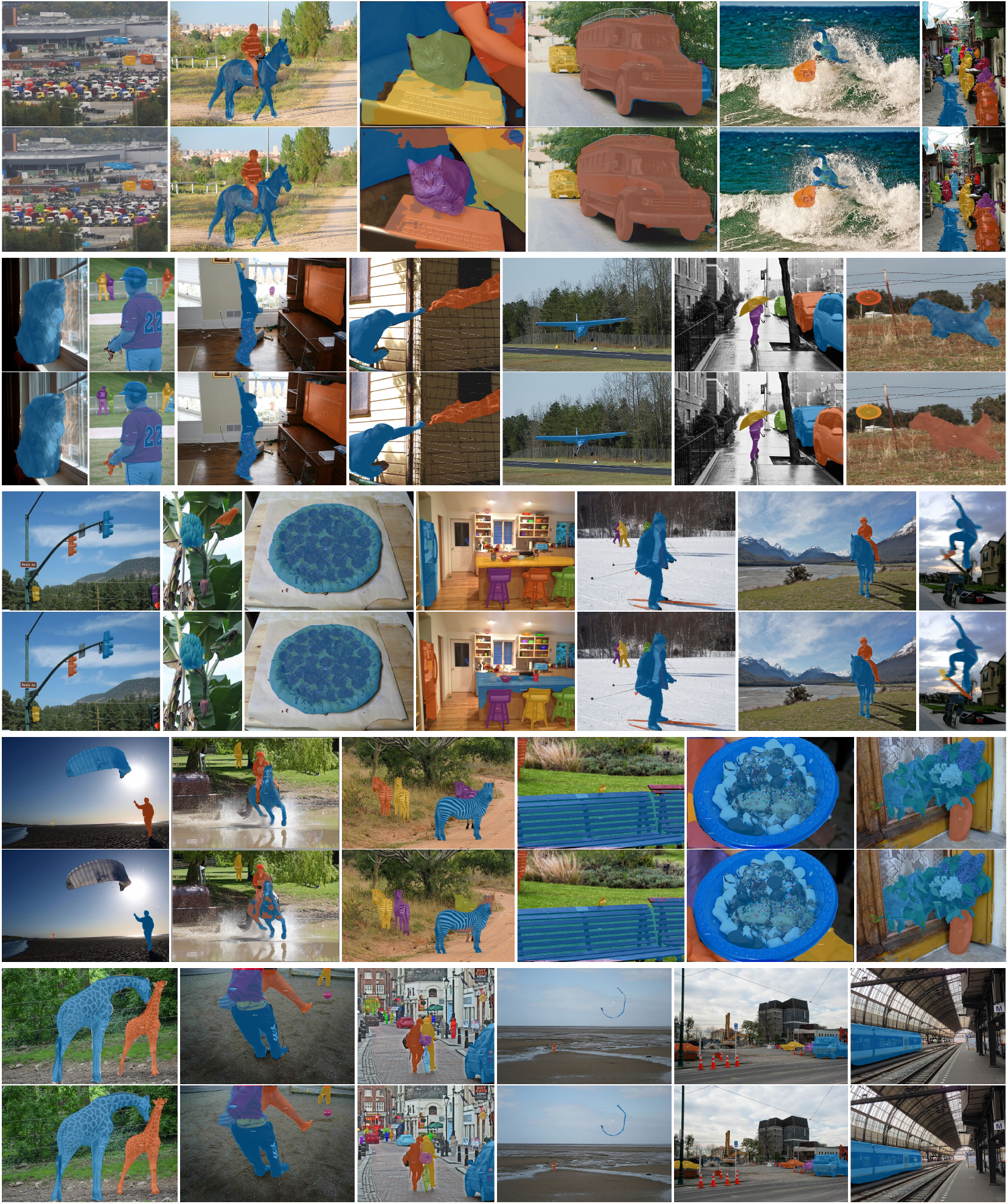}
\caption{More results of Mask R-CNN~\cite{He2017} (top row per set) and \tmask (bottom row per set) on the last $65$ \valset images (continued in Fig.~\ref{fig:qualitative3}). These models use a ResNet-101-FPN backbone and obtain 38.3 and 37.1 AP, on \testset, respectively. Visually, \tmask gives sharper masks compared to Mask R-CNN although its AP is 1 point lower. Best viewed in a digital format with zoom.}
\label{fig:qualitative2}
\end{figure*}

\begin{figure*}\center
 \includegraphics[width=1\linewidth]{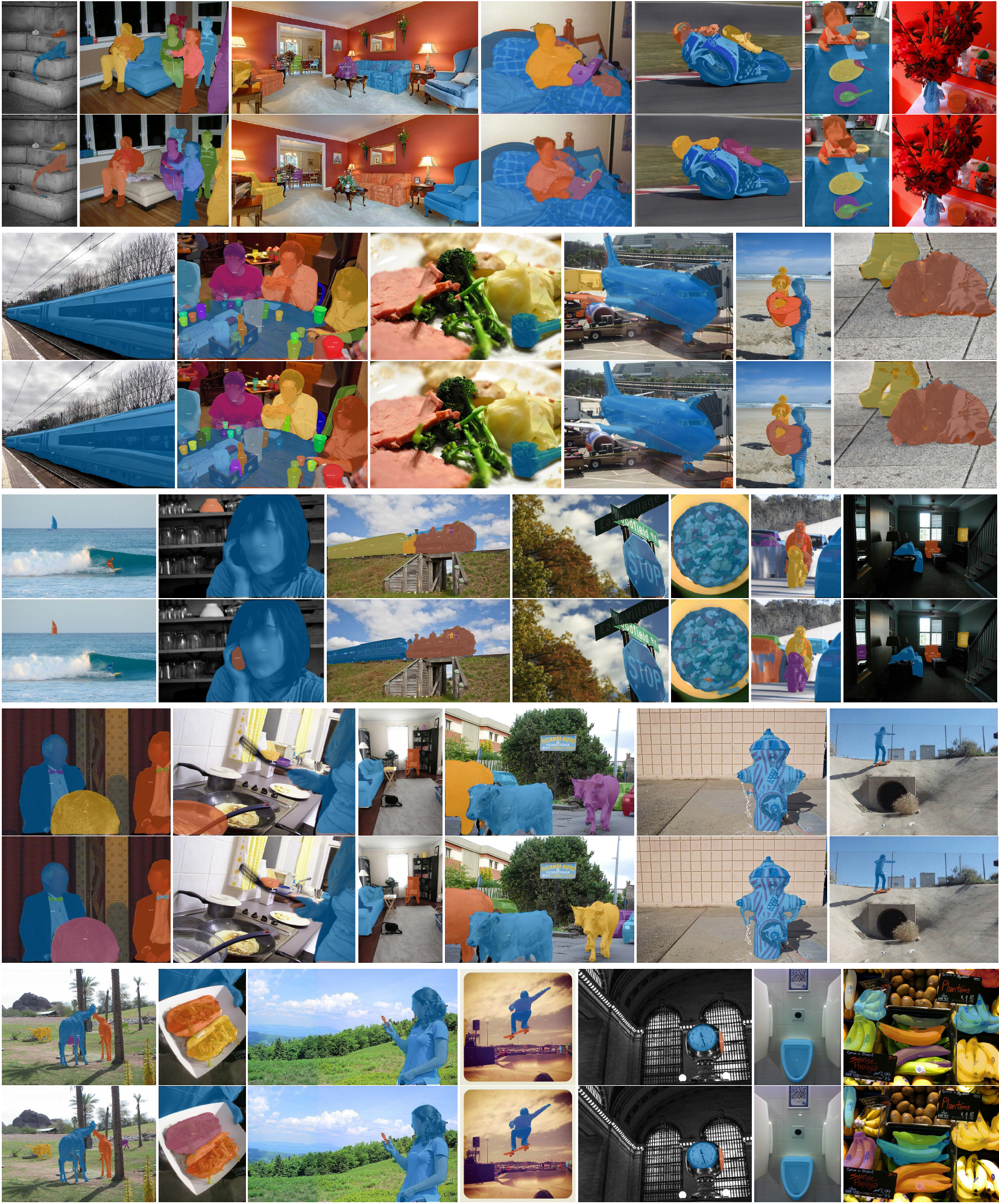}
\caption{More results of Mask R-CNN~\cite{He2017} (top row per set) and \tmask (bottom row per set) continued from Fig.~\ref{fig:qualitative2}.}
\label{fig:qualitative3}
\end{figure*}

We show more results in Figs.~\ref{fig:qualitative2} and \ref{fig:qualitative3}. For these, and all visualizations in the main text, we display all detections that have a \emph{calibrated} score $\ge$0.6. We use a simple calibration that maps uncalibrated detector scores to precision values: for each model and for each category, we compute its precision-recall (PR) curve on \valset. As a PR curve is parameterized by score, we can map an uncalibrated score for the detector-category pair to its corresponding precision value. Score-to-precision calibration enables a fair visual comparison between methods using a fixed threshold.

\newpage
{\setstretch{.99}\small\bibliographystyle{ieee_fullname}\bibliography{tensormask}}

\begin{thebibliography}{10}\itemsep=-1pt

\bibitem{Arbelaez2014}
Pablo Arbel{\'a}ez, Jordi Pont-Tuset, Jonathan~T Barron, Ferran Marques, and
  Jitendra Malik.
\newblock Multiscale combinatorial grouping.
\newblock In {\em CVPR}, 2014.

\bibitem{Arnab2017}
Anurag Arnab and Philip~HS Torr.
\newblock Pixelwise instance segmentation with a dynamically instantiated
  network.
\newblock In {\em CVPR}, 2017.

\bibitem{Bai2017}
Min Bai and Raquel Urtasun.
\newblock Deep watershed transform for instance segmentation.
\newblock In {\em CVPR}, 2017.

\bibitem{Chen2019}
Kai Chen, Jiangmiao Pang, Jiaqi Wang, Yu Xiong, Xiaoxiao Li, Shuyang Sun,
  Wansen Feng, Ziwei Liu, Jianping Shi, Wanli Ouyang, et~al.
\newblock Hybrid task cascade for instance segmentation.
\newblock {\em arXiv:1901.07518}, 2019.

\bibitem{Chen2015}
Liang-Chieh Chen, George Papandreou, Iasonas Kokkinos, Kevin Murphy, and Alan~L
  Yuille.
\newblock Semantic image segmentation with deep convolutional nets and fully
  connected crfs.
\newblock In {\em ICLR}, 2015.

\bibitem{Cordts2016}
Marius Cordts, Mohamed Omran, Sebastian Ramos, Timo Rehfeld, Markus Enzweiler,
  Rodrigo Benenson, Uwe Franke, Stefan Roth, and Bernt Schiele.
\newblock The cityscapes dataset for semantic urban scene understanding.
\newblock In {\em CVPR}, 2016.

\bibitem{Dai2016}
Jifeng Dai, Kaiming He, Yi Li, Shaoqing Ren, and Jian Sun.
\newblock Instance-sensitive fully convolutional networks.
\newblock In {\em ECCV}, 2016.

\bibitem{Dai2016b}
Jifeng Dai, Kaiming He, and Jian Sun.
\newblock Instance-aware semantic segmentation via multi-task network cascades.
\newblock In {\em CVPR}, 2016.

\bibitem{Dollar2009}
Piotr Doll{\'a}r, Zhuowen Tu, Pietro Perona, and Serge Belongie.
\newblock Integral channel features.
\newblock In {\em BMVC}, 2009.

\bibitem{Felzenszwalb2010}
Pedro~F Felzenszwalb, Ross~B Girshick, David McAllester, and Deva Ramanan.
\newblock Object detection with discriminatively trained part-based models.
\newblock {\em PAMI}, 2010.

\bibitem{Girshick2015}
Ross Girshick.
\newblock {Fast R-CNN}.
\newblock In {\em ICCV}, 2015.

\bibitem{Girshick2014}
Ross Girshick, Jeff Donahue, Trevor Darrell, and Jitendra Malik.
\newblock Rich feature hierarchies for accurate object detection and semantic
  segmentation.
\newblock In {\em CVPR}, 2014.

\bibitem{Detectron2018}
Ross Girshick, Ilija Radosavovic, Georgia Gkioxari, Piotr Doll\'{a}r, and
  Kaiming He.
\newblock Detectron.
\newblock \url{https://github.com/facebookresearch/detectron}, 2018.

\bibitem{Goyal2017}
Priya Goyal, Piotr Doll{\'a}r, Ross Girshick, Pieter Noordhuis, Lukasz
  Wesolowski, Aapo Kyrola, Andrew Tulloch, Yangqing Jia, and Kaiming He.
\newblock Accurate, large minibatch {SGD}: Training {ImageNet} in 1 hour.
\newblock {\em arXiv:1706.02677}, 2017.

\bibitem{Hariharan2014}
Bharath Hariharan, Pablo Arbel{\'a}ez, Ross Girshick, and Jitendra Malik.
\newblock Simultaneous detection and segmentation.
\newblock In {\em ECCV}, 2014.

\bibitem{He2018}
Kaiming He, Ross Girshick, and Piotr Doll{\'a}r.
\newblock Rethinking imagenet pre-training.
\newblock {\em arXiv:1811.08883}, 2018.

\bibitem{He2017}
Kaiming He, Georgia Gkioxari, Piotr Doll{\'a}r, and Ross Girshick.
\newblock {Mask R-CNN}.
\newblock In {\em ICCV}, 2017.

\bibitem{He2016}
Kaiming He, Xiangyu Zhang, Shaoqing Ren, and Jian Sun.
\newblock Deep residual learning for image recognition.
\newblock In {\em CVPR}, 2016.

\bibitem{Kirillov2017}
Alexander Kirillov, Evgeny Levinkov, Bjoern Andres, Bogdan Savchynskyy, and
  Carsten Rother.
\newblock Instancecut: from edges to instances with multicut.
\newblock In {\em CVPR}, 2017.

\bibitem{LeCun1989}
Yann LeCun, Bernhard Boser, John~S Denker, Donnie Henderson, Richard~E Howard,
  Wayne Hubbard, and Lawrence~D Jackel.
\newblock Backpropagation applied to handwritten zip code recognition.
\newblock {\em Neural computation}, 1989.

\bibitem{Li2017}
Yi Li, Haozhi Qi, Jifeng Dai, Xiangyang Ji, and Yichen Wei.
\newblock Fully convolutional instance-aware semantic segmentation.
\newblock In {\em CVPR}, 2017.

\bibitem{Lin2017}
Tsung-Yi Lin, Piotr Doll{\'a}r, Ross Girshick, Kaiming He, Bharath Hariharan,
  and Serge Belongie.
\newblock Feature pyramid networks for object detection.
\newblock In {\em CVPR}, 2017.

\bibitem{Lin2017a}
Tsung-Yi Lin, Priya Goyal, Ross Girshick, Kaiming He, and Piotr Doll{\'a}r.
\newblock Focal loss for dense object detection.
\newblock In {\em ICCV}, 2017.

\bibitem{Lin2014}
Tsung-Yi Lin, Michael Maire, Serge Belongie, James Hays, Pietro Perona, Deva
  Ramanan, Piotr Doll{\'a}r, and C~Lawrence Zitnick.
\newblock {Microsoft COCO: Common objects in context}.
\newblock In {\em ECCV}, 2014.

\bibitem{Liu2017a}
Shu Liu, Jiaya Jia, Sanja Fidler, and Raquel Urtasun.
\newblock {SGN}: Sequential grouping networks for instance segmentation.
\newblock In {\em ICCV}, 2017.

\bibitem{Liu2018}
Shu Liu, Lu Qi, Haifang Qin, Jianping Shi, and Jiaya Jia.
\newblock Path aggregation network for instance segmentation.
\newblock In {\em CVPR}, 2018.

\bibitem{Liu2016}
Wei Liu, Dragomir Anguelov, Dumitru Erhan, Christian Szegedy, Scott Reed,
  Cheng-Yang Fu, and Alexander~C Berg.
\newblock {SSD}: Single shot multibox detector.
\newblock In {\em ECCV}, 2016.

\bibitem{Long2015}
Jonathan Long, Evan Shelhamer, and Trevor Darrell.
\newblock Fully convolutional networks for semantic segmentation.
\newblock In {\em CVPR}, 2015.

\bibitem{Neuhold2017}
Gerhard Neuhold, Tobias Ollmann, Samuel Rota~Bulo, and Peter Kontschieder.
\newblock The mapillary vistas dataset for semantic understanding of street
  scenes.
\newblock In {\em ICCV}, 2017.

\bibitem{Peng2018}
Chao Peng, Tete Xiao, Zeming Li, Yuning Jiang, Xiangyu Zhang, Kai Jia, Gang Yu,
  and Jian Sun.
\newblock {MegDet}: A large mini-batch object detector.
\newblock In {\em CVPR}, 2018.

\bibitem{Pinheiro2015}
Pedro Pinheiro, Ronan Collobert, and Piotr Doll{\'a}r.
\newblock Learning to segment object candidates.
\newblock In {\em NIPS}, 2015.

\bibitem{Pinheiro2016}
Pedro Pinheiro, Tsung-Yi Lin, Ronan Collobert, and Piotr Doll{\'a}r.
\newblock Learning to refine object segments.
\newblock In {\em ECCV}, 2016.

\bibitem{Redmon2017}
Joseph Redmon and Ali Farhadi.
\newblock {YOLO9000}: better, faster, stronger.
\newblock In {\em CVPR}, 2017.

\bibitem{Ren2015}
Shaoqing Ren, Kaiming He, Ross Girshick, and Jian Sun.
\newblock {Faster R-CNN}: Towards real-time object detection with region
  proposal networks.
\newblock In {\em NIPS}, 2015.

\bibitem{Rush2019}
Alexander Rush.
\newblock Tensor considered harmful.
\newblock 2019.

\bibitem{Lecun94}
R. Vaillant, C. Monrocq, and Y. LeCun.
\newblock Original approach for the localisation of objects in images.
\newblock {\em IEE Proc. on Vision, Image, and Signal Processing}, 1994.

\bibitem{Sande2011}
Koen~EA van~de Sande, Jasper~RR Uijlings, Theo Gevers, and Arnold~WM Smeulders.
\newblock Segmentation as selective search for object recognition.
\newblock In {\em ICCV}, 2011.

\bibitem{Viola2001}
Paul Viola and Michael Jones.
\newblock Rapid object detection using a boosted cascade of simple features.
\newblock In {\em CVPR}, 2001.

\bibitem{Zagoruyko2016}
Sergey Zagoruyko, Adam Lerer, Tsung-Yi Lin, Pedro Pinheiro, Sam Gross, Soumith
  Chintala, and Piotr Doll{\'a}r.
\newblock A multipath network for object detection.
\newblock In {\em BMVC}, 2016.

\end{thebibliography}

\end{document}